\documentclass{article}

\PassOptionsToPackage{numbers, compress}{natbib}

     \usepackage[preprint]{neurips_2019}

\usepackage[utf8]{inputenc} 
\usepackage[T1]{fontenc}    
\usepackage{hyperref}       
\usepackage{url}            
\usepackage{booktabs}       
\usepackage{amsfonts}       
\usepackage{mathtools} 
\usepackage{amssymb}  
\usepackage{wrapfig}
\usepackage{nicefrac}       
\usepackage{microtype}      
\usepackage{graphics} 
\usepackage{epsfig} 
\usepackage{xcolor}
\usepackage{subcaption}
\usepackage[normalem]{ulem}

\title{Probabilistic Category-Level Pose Estimation via Segmentation and Predicted-Shape Priors}

\author{%
  Benjamin Burchfiel \\
  Department of Computer Science \\
  Duke University \\
  Durham, NC 27708 \\
  \texttt{bcburch@cs.duke.edu} \\
  \And
  George Konidaris \\
  Department of Computer Science \\
  Brown University \\
  Providence, RI 02912 \\
  \texttt{gdk@cs.brown.edu} \\
}

\begin{document}

\maketitle

\begin{abstract}
We introduce a new method for category-level pose estimation which produces a distribution over predicted poses by integrating 3D shape estimates from a generative object model with segmentation information. Given an input depth-image of an object, our variable-time method uses a mixture density network architecture to produce a multi-modal distribution over 3DOF poses; this distribution is then combined with a prior probability encouraging silhouette agreement between the observed input and predicted object pose. Our approach significantly outperforms the current state-of-the-art in category-level 3DOF pose estimation---which outputs a point estimate and does not explicitly incorporate shape and segmentation information---as measured on the Pix3D and ShapeNet datasets.

\end{abstract}

\section{Introduction}
A critical task in many robotics and computer vision applications is determining an observed object's 3D pose. While some approaches attempt to estimate only a subset of 3D pose parameters \cite{sun2018pix3d, Su_2015_ICCV}, general solutions must produce a full estimate of the object's 3D orientation. Broadly speaking, full 3D pose estimation falls into two categories: instance-level and category-level. While the instance-level case assumes availability of a full 3D ground-truth model of the object, category-level pose estimation must deal with a novel object for which no ground-truth model is available. In the category-level case, it is common to define a canonical pose---a default pose that is well defined with respect to a particular class of objects. For instance, the canonical pose of a car might be with the front bumper directly facing the camera and all wheels on the ground plane. Category-level pose estimation approaches generally learn to transform an object into these canonical poses directly from training data (often implicitly).

Recent work in category-level pose estimation includes both discrete pose estimators \cite{sun2018pix3d, Su_2015_ICCV, NIPS2012_4562}, which treat pose estimation as a classification task by predicting a single pose bucket (generally with width around 15 degrees), and continuous methods which directly output pose parameters \cite{beo2017, hbeo2018}. Notably, while some of these approaches estimate both object pose and 3D object shape from a single input depth \cite{hbeo2018} or RGB image \cite{sun2018pix3d}, they fail to explicitly verify that their estimates of shape and pose are consistent with the observed depth or RGB input.

We propose a novel pose estimation approach that leverages a generative model of 3D shape to verify that predicted pose matches the segmented input. Our method predicts a distribution over possible object poses from a single segmented depth-image, and constructs a pose consistency prior based on agreement between the predicted object silhouette and input image. We modify an existing generative object model, HBEOs \cite{hbeo2018}, to produce a multimodal distribution over poses instead of a point-estimate, and develop an efficient 2D pose-consistency score by projecting shape-pose estimates back into the input image. We use this consistency score as a prior probability over poses and provide a sampling-based approximation to the \textit{maximum a posteriori} estimate of 3D pose. We evaluate our approach empirically on several thousand 3D objects across three classes from the ShapeNet dataset \cite{shapenet2015} as well as roughly three thousand chairs from the Pix3D dataset \cite{sun2018pix3d}. Our results show a dramatic decrease in gross pose error ($\epsilon > 15^\circ$) compared to previous state-of-the-art and a significant performance contribution from both the density prediction and input-verification components. Furthermore, by dynamically varying the number of samples used to produce a pose-estimate, our approach becomes an any-time method; while it significantly outperforms existing methods with only a few samples, it produces increasingly high-quality pose estimates when given a longer budget.

\section{Background}
Single-instance pose estimation has been widely studied and can be largely broken down into optimization-based and learning-based approaches. Optimization approaches formulate pose estimation as the alignment of an object's known model to observed input; these methods, which include work such as Iterative Closest Points \cite{ICP} and Bingham-Distribution filters \cite{Glover-RSS-11}, attempt to minimize error between object models and observations. More recently, learning-based approaches---which attempt to directly predict object pose---have begun to outperform optimization-based methods due to their robustness to local minima and ability to learn highly discriminative features in input space \cite{tulsiani2015viewpoints, wohlhart2015learning, deep_im, pose_cnn}.

There has been less work on category-level pose estimation---predicting the pose of novel objects \cite{sun2018pix3d, Su_2015_ICCV, NIPS2012_4562, beo2017, hbeo2018} given a canonical notion of pose for each object category. While RGB approaches such as \citet{sun2018pix3d} still tend to be limited to predicting pose in course bins (of roughly 15 degrees of width), a method called HBEOs \cite{hbeo2018} directly predicts an object's continuous 3D orientation from a single input depth-image. The pose-estimation portion of HBEOs is still not reliable enough for robust robot interaction with objects, however, routinely producing pose estimates with errors of over fifteen degrees.


\subsection{Hybrid Bayesian Eigenobjects}

HBEOs are a generative representation of 3D objects; they explicitly model a low-dimensional object-space and are capable of representing novel and partially observed objects from previously encountered categories. HBEOs take as input a segmented depth-image of a novel observed object and estimates the object's 3D shape---in voxel form---along with its class and 3DOF orientation \cite{hbeo2018}. Similarly to other methods that complete 3D objects, HBEOs are trained from a set of aligned 3D meshes. Each mesh is voxelized and each training voxel object of size $m\times n\times o$ is reshaped to produce a size $d=m\cdot n\cdot o$ vector. A Bayesian variant of Principal Component Analysis \cite{pca}, VBPCA \cite{vbpca}, is used to learn a loose low-dimensional subspace that captures the shape variation in each class of training objects. We denote the learned basis-matrix and mean-vector for class $i$ as $\mathbf{W_i}$ and $\boldsymbol{\mu_i}$, respectively; from these class-specific subspaces, a single shared subspace is created by finding a basis spanning each subspace:
\begin{equation*}
\mathbf{W} = [\mathbf{W_1},...,\mathbf{W_m}, \boldsymbol{\mu_1},...,\boldsymbol{\mu_m}].
    \label{eq:bigSubspace}
\end{equation*}
To ensure this basis is maximally compact, singular value decomposition is used to find a minimal orthonormal representation of $\mathbf{W}$. The projection of a voxel object in vector form, $\mathbf{o}$, into this subspace is
\begin{equation}
\mathbf{o}'=\mathbf{W}^T\mathbf{o}
\label{eq:project_complete}
\end{equation}
and any point in this subspace can be mapped back into voxel-space via
\begin{equation}
\mathbf{\hat{o}}= \mathbf{W}\mathbf{o}'.
\label{eq:reconst}
\end{equation}

HBEOs then train a CNN to directly estimate an object's projection onto this space, $\mathbf{o}'$, given a single depth-view of that object. Because $\mathbf{W}$ has been learned during training, at inference time the HBEO prediction $\hat{\mathbf{o}}'$ is sufficient to estimate full 3D geometry using equation \ref{eq:reconst}. HBEOs also simultaneously predict a probability distribution over object classes and a point-estimate of pose in the form of three axis-angle pose parameters. 

\subsection{Mixture Density Networks}
Mixture Density Networks (MDNs) were first proposed by \citet{mdns} as a method for predicting distributions, rather than point-estimates, when using neural networks for regression. MDNs explicitly estimate a distribution that explains the training data; let $x$ be inputs to the model and $y$ be the desired regression target. A conventional regression network predicts $\hat{y}$ directly from $x$ while an MDN predicts a conditional density function $P(y | x)$.

MDNs model $P(y | x)$ as a mixture of parameterized distributions; a common choice being the multivariate Gaussian mixture model (MV-GMM) with probability density function:
\begin{equation}
    p({\boldsymbol {y |\theta}})=\sum _{i=1}^{c}{\pi_{i}}{\mathcal {N}}(y | {\boldsymbol {{\mu _{i}},{\Sigma _{i}}}}),
    \label{eq:pose_likelihood}
\end{equation}
where $\theta_i=\{\boldsymbol {\pi_i,{\mu _{i}},{\Sigma _{i}}}\}$ is the mixing coefficient, mean, and covariance of component $i$. The network predicts these mixture parameters by learning the function $\theta = f(x)$ while a conventional regression network learns $y=f(x)$. As a result, if $y\in\mathbb{R}^n$, a network that directly predicts $y$ would have size $n$ output while an MV-GMM-MDN would produce output of size $c(n^2+n+1)$, where $c$ is the number of mixture components. To reduce the output size of MDNs, it is common to assume a diagonal covariance for each component, in which case the output size becomes $c(2n+1)$. During training, each gradient update seeks to minimize the negative log-likelihood of observed data:
\begin{equation}
    loss(y , \theta) = -ln{\sum _{i=1}^{c}{\pi_{i}}{\mathcal {N}}(y | {\boldsymbol {{\mu _{i}},{\Sigma _{i}}}})}.
    \label{eq:mdn_loss}
\end{equation}

MDN networks have some significant advantages over direct regression including non-unimodality and measurable uncertainty. While direct regression learns to predict the conditional mean of the data, MDNs can learn to model the shape of the underlying distribution. In contexts where the target is ill-posed (i.e. there are multiple valid mappings from $x$ to $y$), the mean over good solutions may not actually be a reasonable solution itself.
\begin{wrapfigure}{r}{0.4\textwidth}
  \begin{center}
    \vspace{-5mm}
    \includegraphics[width=0.4\textwidth]{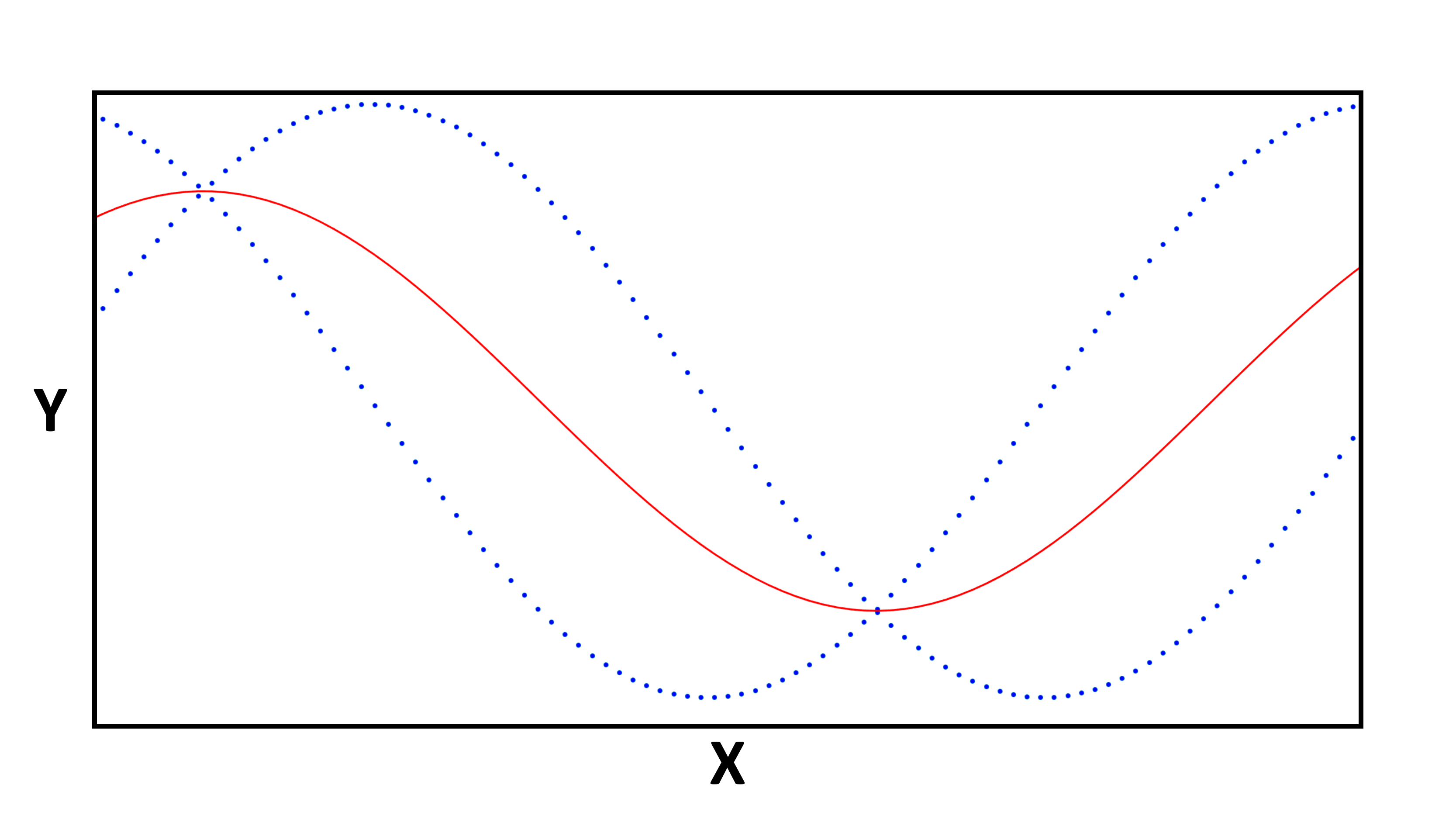}
  \end{center}
  \caption{Blue: training points. Red: $f(x)$ learned through direct regression.}
  \vspace{-20mm}
  \label{fig:multimode_example}
\end{wrapfigure}
Figure \ref{fig:multimode_example} illustrates this in a simple univariate scenario; the function learned by direct regression with a least-squares loss is unable to accurately represent the underlying data. MDNs also allow multiple solutions to be sampled and provide a measure of confidence for each sample via the predicted conditional PDF. This explicit likelihood estimate provides a natural avenue for incorporating prior information into estimates in a straightforward and principled way.

\section{Predicting Pose Posteriors}
We improve pose estimation in two primary ways: 1) predicting multi-modal pose distributions instead of point-estimates and 2) incorporating a prior probability dependent on agreement between predicted pose, shape, and the input depth-image. We construct a distribution-predicting network using the MDN formulation and show how approximate maximum likelihood and \textit{maximum a posteriori} estimates may be obtained via sampling from its output.

\subsection{Using Predicted-Shape Priors for Predicting Pose Distributions}
In order to incorporate pose priors, we require a probabilistic estimate of pose distribution conditioned on input observation. To obtain this, we modify the HBEO architecture \cite{hbeo2018} to estimate a pose distribution by converting the final layer of the network to a multivariate Gaussian MDN layer. Let $z_i$ be the sum of input signals to output neuron $i$ and assume predicted components have diagonal covariance. Each covariance value is estimated using a translated exponential-linear unit (ELU) at the final layer:
  \[
    \sigma_{ii} = \left\{\begin{array}{lr}
        z_i, & \text{if} x > 0\\
        \alpha~exp(z_i) + \epsilon, & \text{if } x \leq 0\\
        \end{array}\right\}.
  \]
This translated ELU ensures that elements along the diagonal of the predicted covariance matrix will be strictly positive and that the resulting matrix will be positive semidefinite. Component mean parameters are estimated straightforwardly with a linear layer and component mixing coefficients, $\pi\in\Pi$, are estimated with a softmax layer, ensuring $\Pi$ forms a valid discrete probability distribution:
\begin{equation*}
\pi_i=\frac{e^{z_i}}{\Sigma_{j=1}^ce^{z_j}}.
\end{equation*}
We fixed the number of components to $c=5$ and used equation \ref{eq:mdn_loss} to calculate the pose portion of our loss function:
\begin{equation}
    loss_{pose}(y, \theta) = -ln{\sum _{i=1}^{5}{\pi_{i}}}\frac {\exp \left(-{\frac {1}{2}}({\mathbf {y} }-{\boldsymbol {\mu }})^{\mathrm {T} }{\boldsymbol {\Sigma }}^{-1}({\mathbf {y} }-{\boldsymbol {\mu }})\right)}{\sqrt {(2\pi )^{k}|{\boldsymbol {\Sigma }}|}},
    \label{eq:pose_loss}
\end{equation}
where $\theta$ are the distribution parameters predicted by the network for input image $x$ and $y$ is the true pose of the object depicted in $x$. We maintain the same structure used in \citet{hbeo2018} for the lower layers of the network and for the class and shape output layers---softmax and linear layers, respectively.
The full loss for the network is
\begin{equation}
\mathcal{L}(y) = \lambda_{p}loss_{pose}(y) + \lambda_{s}loss_{shape}(y) + \lambda_{c}loss_{class}(y)
\end{equation}
where $loss_{shape}(y)$ is a Euclidean loss over subspace projection coefficients, $loss_{class}(y)$ is the multiclass categorical cross-entropy loss over possible classes, and $\lambda_{p}$, $\lambda_{s}$, $\lambda_{c}$ are weighting coefficients over the pose, shape, and class terms.\footnote{In our experiments, we found that $\lambda_{p} = \lambda_{s} = \lambda_{c} = 1$ yielded good results.} Beyond allowing multiple possible poses to be sampled, HBEO-MDNs are more robust to training data pose noise and object symmetry than HBEOs because they can explicitly model multiple pose modalities. Furthermore, MDNs naturally compensate for the representational discontinuity present in axis-angle formulations of pose. As an example, consider predicting only the z-axis rotation component of an object's pose. If the true pose z-component is $\pi$, and target poses are are in the range of $(-\pi, \pi]$, then the HBEO network would receive a small loss for predicting $p_z=\pi-\epsilon$ and a large loss for predicting $p_z=\pi+\epsilon$, despite the fact that both predictions are close to the true pose. While other loss functions or pose representations may alleviate this particular issue, they do so at the expense of introducing problems such as double coverage, causing the network's prediction target to no longer be well defined. By comparison, the HBEO-MDN approach suffers from none of these issues and can model object symmetry and discontinuities in prediction space explicitly by predicting multi-modal distributions over pose.

\subsection{Pose Priors from Shape and Segmentation}
While generative models that predict an object's 3D shape exist, those that also estimate object pose do not explicitly verify that these predictions are consistent with observed depth input. While the shape estimate produced by such models is noisy, there is still valuable information to be obtained from such a verification. Let $D$ be a segmented depth-image input to such a model, $\mathbf{\hat{o}}$ be the predicted shape of the object present in $D$, and $\mathbf{\hat{R}(\mathbf{\hat{o}})}$ be the estimated 3DOF rotation transforming $\mathbf{\hat{o}}$ from canonical pose to the pose depicted in $D$. Assuming known depth camera intrinsic parameters, we can project the estimated shape and pose of the object back into a 2D depth-image via $\hat{D_R} = f(\mathbf{\hat{R}(\mathbf{\hat{o}})})$ where the projection function $f(x)$ simulates a depth camera. Intuitively, if the shape and pose of the observed object are correctly estimated, and the segmentation and camera intrinsics are accurate, then $\Delta_D = ||D -\hat{D_R}||=0$ while errors in these estimates will result in a discrepancy between the predicted and observed depth-images. As prior work has shown pose to be the least reliable part of the pipeline \cite{beo2017, hbeo2018, sun2018pix3d}, we assume that the error in $\hat{R}$ will generally dominate error in the other portions of the process and thus employ $\Delta_D$ to refine $\hat{R}$.

\begin{figure*}[ht!]
\centering
\includegraphics[width= 1.0\textwidth]{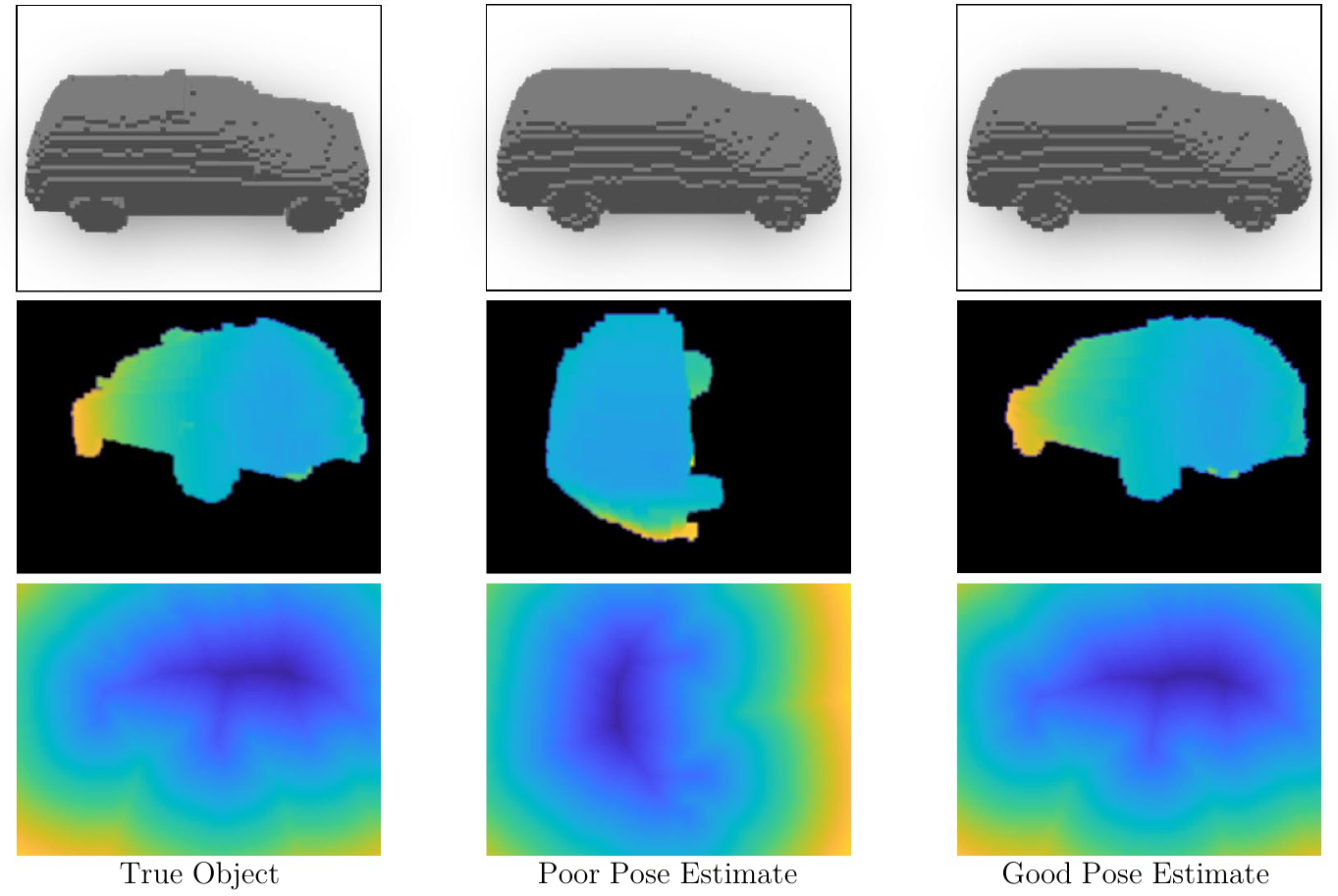}
\caption{Example output of HBEO-MDN net evaluated on a car. The first column denotes the true 3D object (top), observed depth-image (middle) and resulting SDF (bottom) while the second and third columns depict estimated 3D object shape, depth-image, and resulting SDF. Note that the SDF corresponding to an accurate pose-estimate closely matches that of the observed input while the poor estimate does not.}
\label{fig:sdf_example}
\end{figure*}

Let $T=SDF(D)$ be the 2D \emph{signed distance field} \cite{TSDF} calculated from $D$, we define an image-space error score between segmented depth-images as
\begin{equation}
    e_R=||SDF(D) - SDF(\hat{D_R})||_{f}
    \label{eq:depth_error}
\end{equation}
where $SDF(D)$ considers all non-masked depth values to be part of the object and $||\cdot||_{f}$ denotes the Frobenius norm. Figure \ref{fig:sdf_example} illustrates the masked depth input and SDF for an example object. The calculation of error in depth-image-domain has several advantages; because it operates in 2D image space, distance fields are both more efficient to calculate than in the 3D case and better defined because the observed input is partially occluded in 3D but fully observable from the single 2D perspective of the camera. Furthermore, by using the SDF instead of raw depth values, our error gains some robustness to sensor noise and minor errors in predicted shape. To transform this error into a pose prior, we take the quartic-normalized inverse of the score, producing the density function
\begin{equation}
    p_{prior}(\hat{R}) = \frac{1}{e_R^4+\epsilon}.
    \label{eq:pose_prior}
\end{equation}

\subsection{Sampling Pose Estimates}
It is possible to obtain approximate maximum likelihood (MLE) and \textit{maximum a posteriori} (MAP) pose estimates by sampling from the pose distribution induced by the predicted $\theta$. Let $\mathbf{R}$ denote the set of $n$ pose estimates sampled the HBEO-MDN network and $R_i\in \mathbf{R}$ be a single sampled pose. From equation \ref{eq:pose_likelihood}, the approximate MLE pose estimate is
\begin{equation}
    \hat{R}_{MLE} = argmax_{R_i\in \mathbf{R}}~~\sum _{i=1}^{c}{\pi_{i}}{\mathcal {N}}(R_i | {\boldsymbol {{\mu _{i}},{\Sigma _{i}}}})
    \label{eq:MLE}
\end{equation}
while incorporating equation \ref{eq:pose_prior} produces an approximate MAP pose estimate of
\begin{equation}
    \hat{R}_{MAP} = argmax_{R_i\in \mathbf{R}}~~\frac{1}{e_{R_{i}}^4+\epsilon}\sum _{i=1}^{c}{\pi_{i}}{\mathcal {N}}(R_i | {\boldsymbol {{\mu _{i}},{\Sigma _{i}}}}).
    \label{eq:MAP}
\end{equation}
As $n\to\infty$, equations \ref{eq:MLE} and \ref{eq:MAP} approach the true MLE and MAP pose estimates, respectively. As a result, HBEO-MDN is a variable-time method for pose estimation, with prediction accuracy improving as computation time increases.

\section{Experimental Evaluation}
We evaluated our approach via ablation on three datasets consisting of cars, planes, and couches taken from ShapeNet \cite{shapenet2015} for a total of $6659$ training objects and $3098$ test objects. We also compared to two RGB-based approaches on the standard Pix3D chair dataset. Depth-based approaches were provided with segmented depth-image input and RGB-based approaches were given tight bounding boxes around objects; in the wild, these segmentations could be estimated using dense semantic segmentation such as MASK-RCNN \cite{he2017mask}. For the ablation experiments, HBEOs \cite{hbeo2018} and HBEO-MDNs were trained for each category of object with $2798$ couches, $2986$ planes, and $875$ cars used. During training, depth-images from random views\footnote{Azimuth and elevation were sampled across the full range of possible angles while roll was sampled from 0-mean Gaussian distribution with 99-percent mass within the range $[-25^\circ$, $25^\circ]$.} were generated for each object for a total of $2.7M$ training images. Evaluation datasets were constructed for each class containing 1500 views from $50$ cars, $2300$ views from $947$ planes, and $2101$ views from $368$ couches. The HBEO and HBEO-MDN models used identical subspaces of size $d=300$ for each object class, predicted size $64^3$ voxel objects, and were trained for $25$ epochs (both models converged at similar rates).\footnote{Models were trained using the Adam optimizer with $\alpha=0.001$ and evaluated on an Nvidia 1080ti GPU.}

We examined two forms of HBEO-MDN, an ablation that used the MLE approximation from equation \ref{eq:MLE} (HBEO-MDN Likelihood) and the full method which uses the posterior approximation defined in equation \ref{eq:MAP} (HBEO-MDN Posterior). The performance of HBEO-MDN Likelihood illustrates the contribution of the MDN portion of our approach while HBEO-MDN Posterior shows the efficacy of explicitly verifying possible solutions against the observed depth-image. To ablate the impact of the generative portion of our model, we also evaluated two baselines, Random Sample + Oracle, which uniformly sampled poses from $SO(3)$ and was provided with an oracle to determine which of the samples poses was closest to ground truth, and Random Sample + SDF Error, which uniformly sampled poses from $SO(3)$ and used equation \ref{eq:depth_error} to select a single pose estimate from these samples.

\begin{figure}
\begin{subfigure}{.5\textwidth}
  \centering
  \includegraphics[width=1\linewidth]{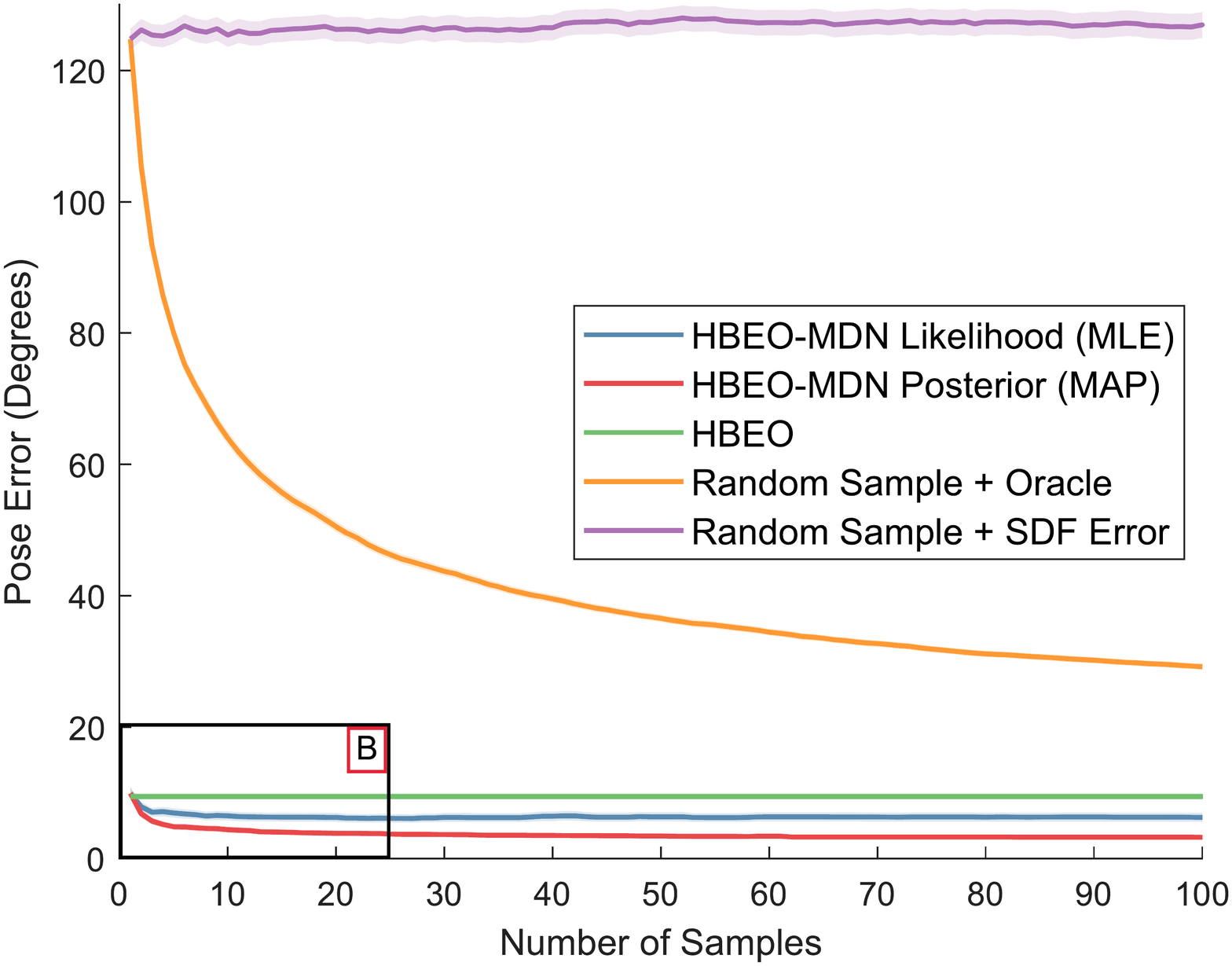}
  \caption{Car Dataset}
  \label{fig:car_plot}
\end{subfigure}%
\begin{subfigure}{.5\textwidth}
  \centering
  \includegraphics[width=1\linewidth]{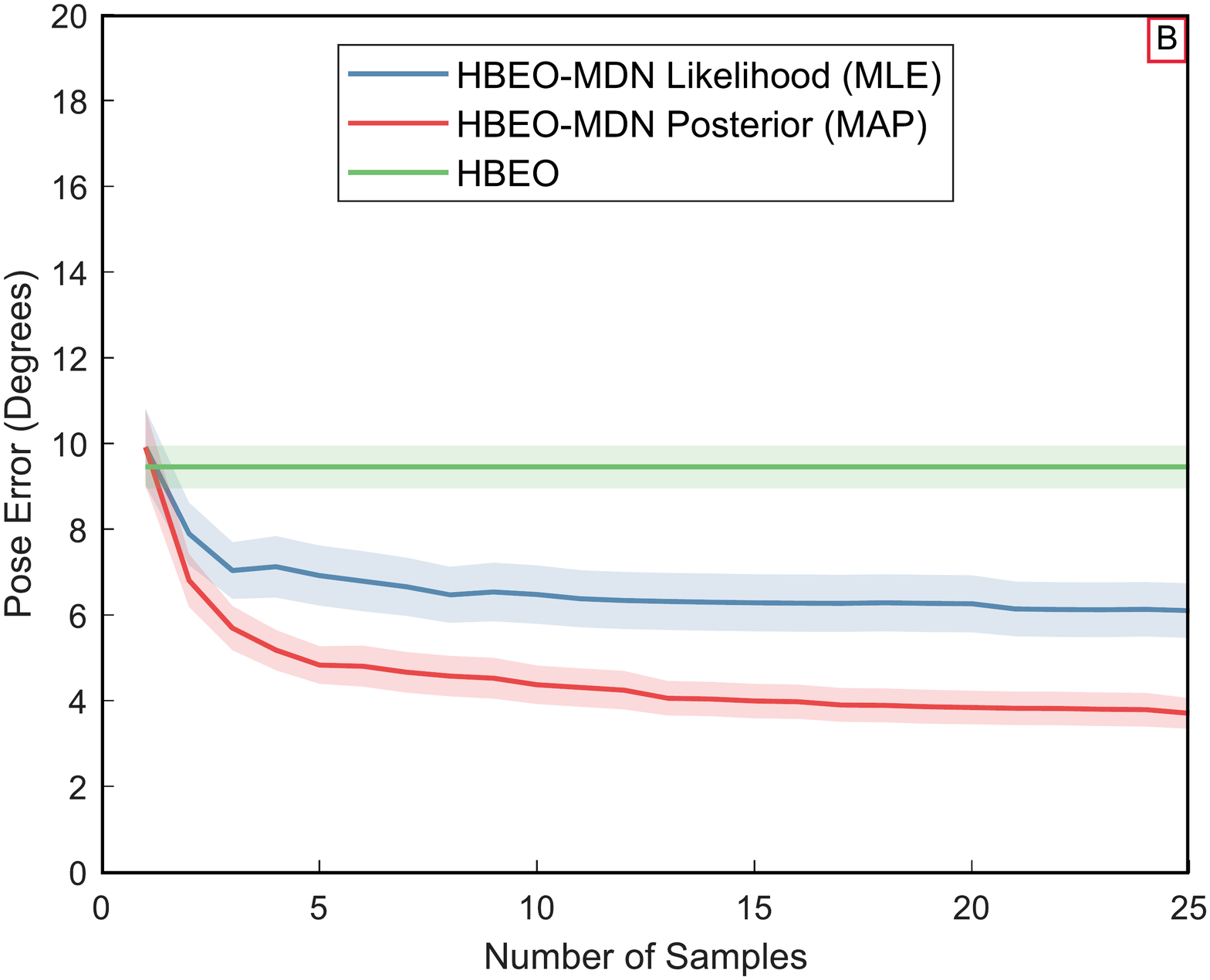}
  \caption{Car Dataset (Enlarged)}
  \label{fig:car_plot_zoom}
\end{subfigure}
\bigskip
\begin{subfigure}{.5\textwidth}
  \centering
  \includegraphics[width=1\linewidth]{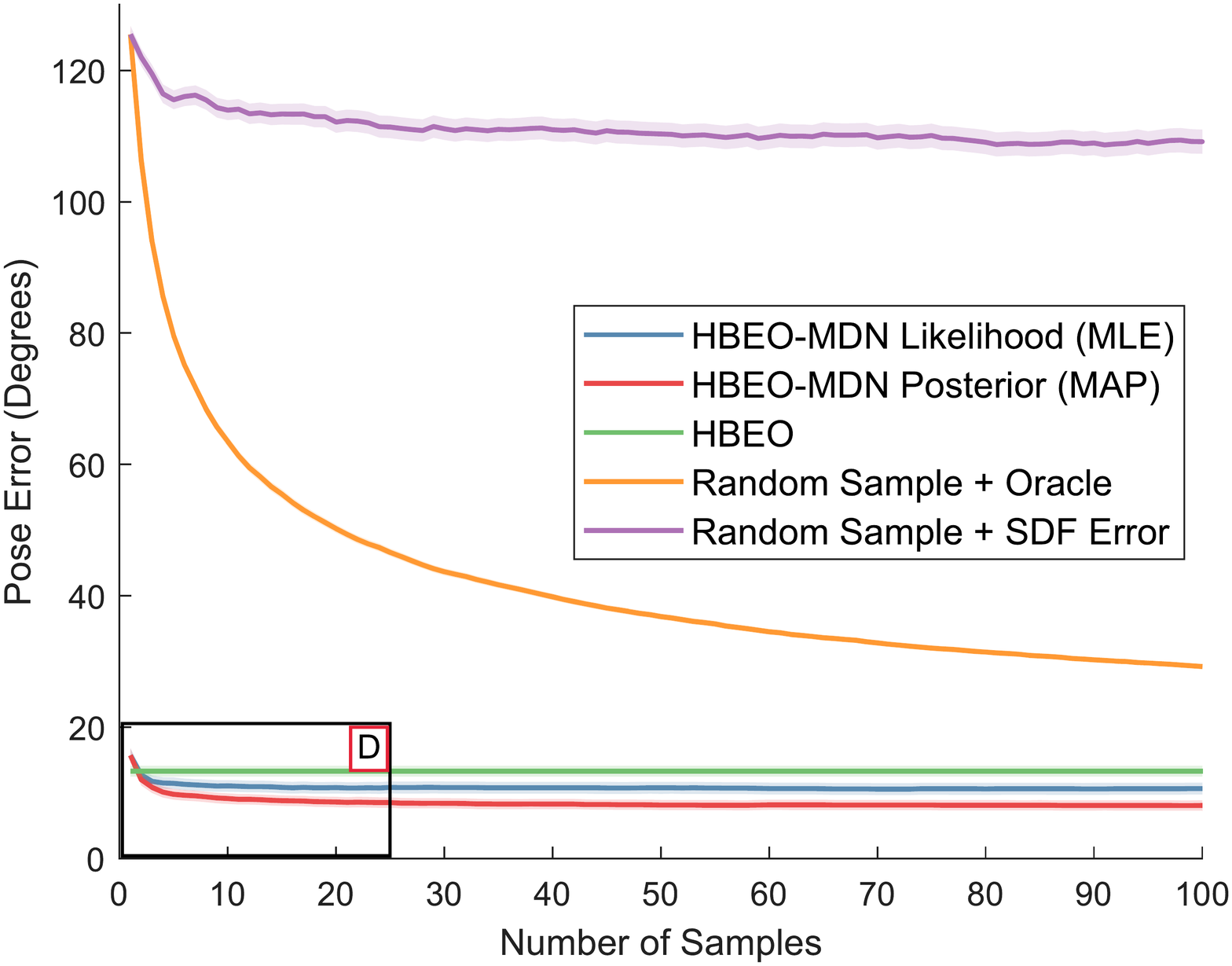}
  \caption{Plane Dataset}
  \label{fig:plane_plot}
\end{subfigure}%
\begin{subfigure}{.5\textwidth}
  \centering
  \includegraphics[width=1\linewidth]{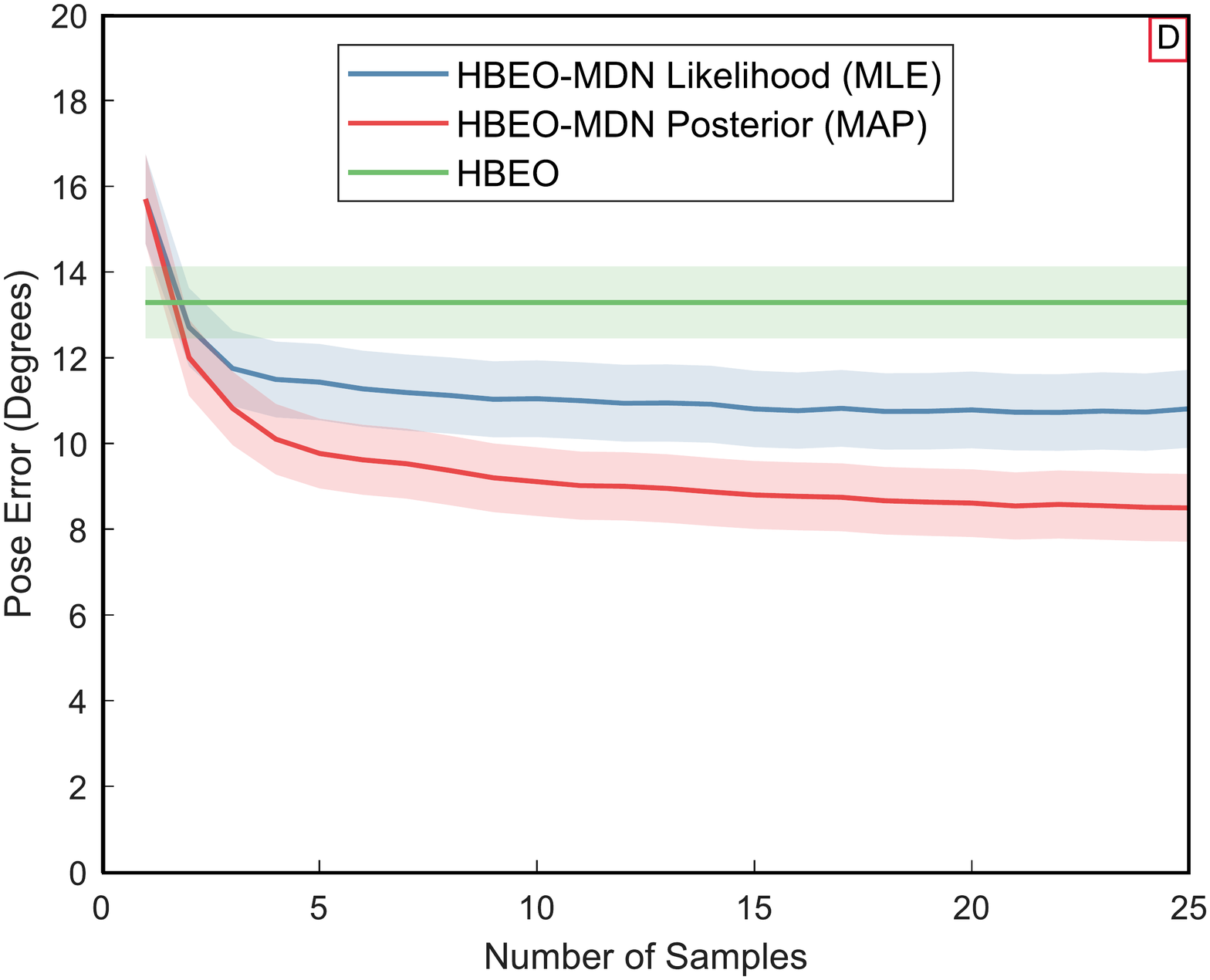}
  \caption{Plane Dataset (Enlarged)}
  \label{fig:plane_plot_zoom}
\end{subfigure}
\bigskip
\begin{subfigure}{.5\textwidth}
  \centering
  \includegraphics[width=1\linewidth]{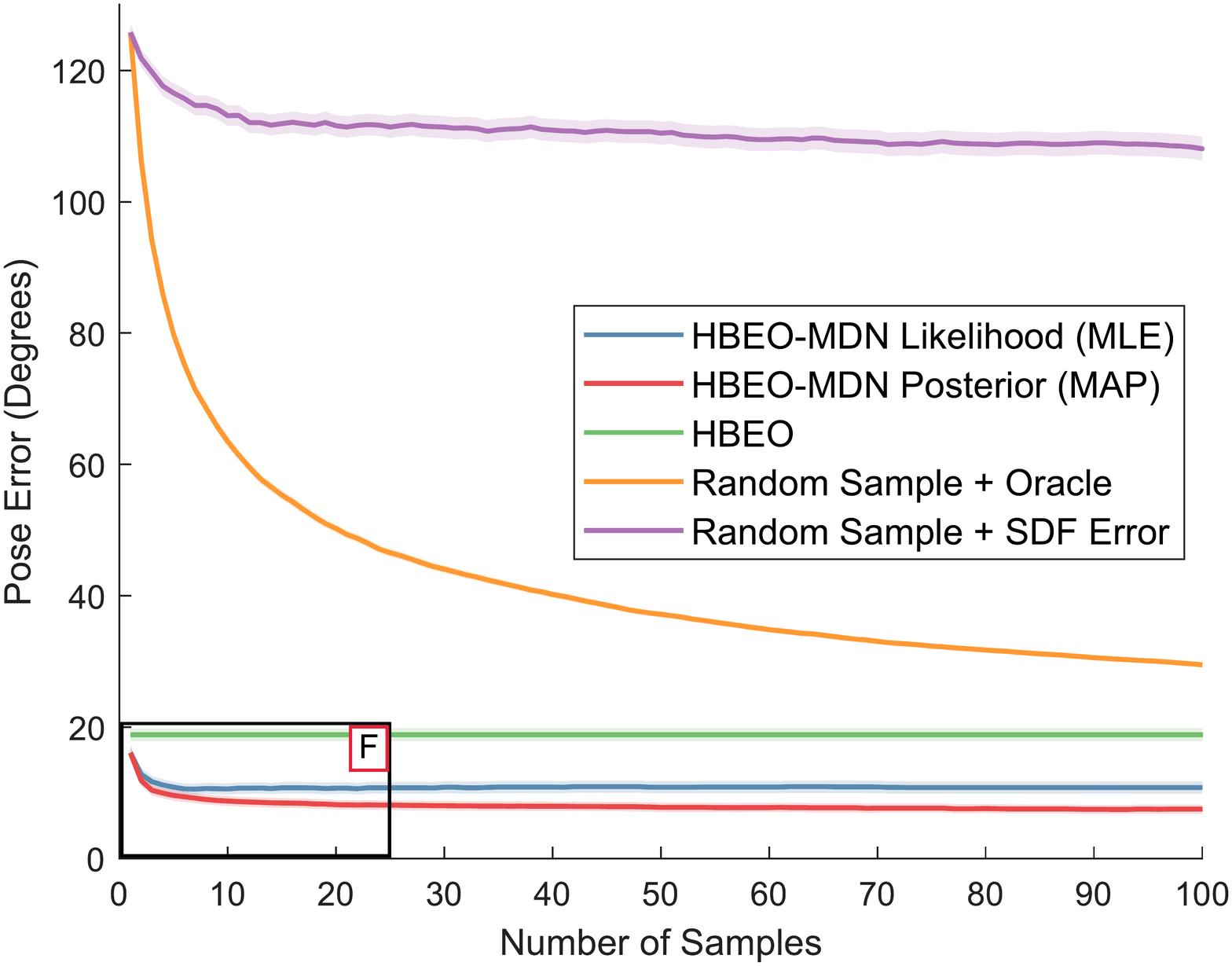}
  \caption{Couch Dataset}
  \label{fig:couch_plot}
\end{subfigure}%
\begin{subfigure}{.5\textwidth}
  \centering
  \includegraphics[width=1\linewidth]{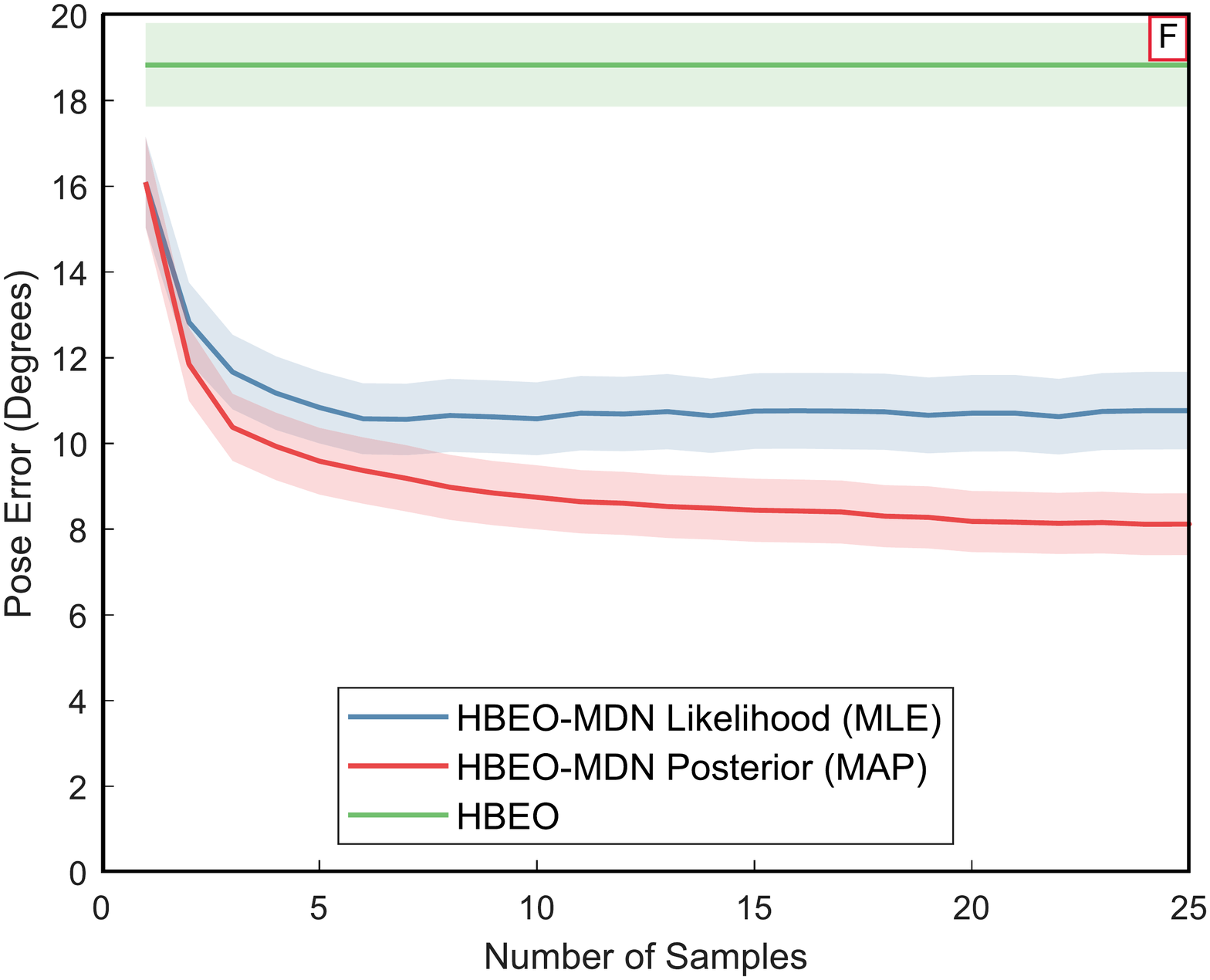}
  \caption{Couch Dataset (Enlarged)}
  \label{fig:couch_plot_zoom}
\end{subfigure}
\caption{Mean pose error per number of samples across three datasets, Cars (top), Planes (middle), and Couches (bottom). HBEO results (green) are constant since HBEOs do not sample solutions. Error ranges indicate the 95 percent confidence estimate of the mean. }
\label{fig:pose_graphs}
\end{figure}

\begin{table}
  \caption{ShapeNet pose estimation performance --- mean error and runtime}
  \label{table:results_table}
  \centering
  \begin{tabular}{llllll}
    \toprule
        \multicolumn{6}{r}{Mean Angular Error (Lower is Better)\hphantom{12345}} \\
    \cmidrule(r){3-5}
    Method & N & Car & Plane & Couch & Runtime \\
    \midrule
    HBEO-MDN Likelihood\hphantom{123} & $5$\hphantom{12} & $6.92^\circ$ & $11.43^\circ$ & $10.84^\circ$ & $0.07$s \\
    HBEO-MDN Likelihood\hphantom{123} & $25$\hphantom{12} & $6.10^\circ$ & $10.81^\circ$ & $10.77^\circ$ & $0.13$s \\
    HBEO-MDN Likelihood\hphantom{123} & $100$\hphantom{12} & $6.26^\circ$ & $10.64^\circ$ & $10.80^\circ$ & $0.36$s \\
    HBEO-MDN Posterior\hphantom{123} & $5$\hphantom{12} & $4.83^\circ$ & $9.77^\circ$ & $9.59^\circ$ & $0.12$s \\
    HBEO-MDN Posterior\hphantom{123} & $25$\hphantom{12} & $3.70^\circ$ & $8.45^\circ$ & $8.12^\circ$ & $0.32$s \\
    HBEO-MDN Posterior\hphantom{123} & $100$\hphantom{12} & $\mathbf{3.23^\circ}$ & $\mathbf{8.06^\circ}$ & $\mathbf{7.50^\circ}$ & $1.06$s \\
    HBEO \cite{hbeo2018} & $N/A$\hphantom{12} & $9.45^\circ$ & $13.29^\circ$ & $18.84^\circ$ & $\mathbf{0.01}$s \\
    \bottomrule
  \end{tabular}
\end{table}
\begin{table}
  \caption{ShapeNet pose estimation performance --- gross-error incidence rate}
  \label{table:results_table2}
  \centering
  \begin{tabular}{lllll}
    \toprule
     \multicolumn{5}{r}{Error $>15^\circ$ (Lower is Better)\hphantom{1}} \\
    \cmidrule(r){3-5}
    Method & N & Car & Plane & Couch \\
    \midrule
    HBEO-MDN Likelihood\hphantom{123} & $5$\hphantom{123} & $3.73~\%$ & $10.87~\%$ & $9.57~\%$ \\
    HBEO-MDN Likelihood\hphantom{123} & $25$\hphantom{123} & $2.67~\%$ & $9.78~\%$ & $9.09~\%$ \\
    HBEO-MDN Likelihood\hphantom{123} & $100$\hphantom{123} & $2.80~\%$ & $9.65~\%$ & $8.95~\%$ \\
    HBEO-MDN Posterior\hphantom{123} & $5$\hphantom{123} & $1.80~\%$ & $8.13~\%$ & $8.09~\%$ \\
    HBEO-MDN Posterior\hphantom{123} & $25$\hphantom{123} & $1.20~\%$ & $6.22~\%$ & $6.38~\%$ \\
    HBEO-MDN Posterior\hphantom{123} & $100$\hphantom{123} & $\mathbf{0.93~\%}$ & $\mathbf{5.78~\%}$ & $\mathbf{5.95~\%}$ \\
    HBEO \cite{hbeo2018} & $N/A$\hphantom{123} & $6.87~\%$ & $13.57~\%$ & $31.27~\%$ \\
    \bottomrule
  \end{tabular}\\
\end{table}

Figure \ref{fig:pose_graphs} gives the performance of the approach across all three ShapeNet evaluation datasets as a function of the number of samples used. Table \ref{table:results_table} contains performance and inference-time at various sampling values while Table \ref{table:results_table2} contains the frequency of large pose errors of at least $15$ degrees. HBEO-MDN Posterior substantially outperformed other approaches, even with a limited sample size.

HBEOs, while having generally inferior performance to both HBEO-MDN varieties, struggled most significantly on the couch dataset. We hypothesize that this is due to the symmetry present in couches when viewed from the side, where only small aspects of the image disambiguate left from right. Because the MDN approaches can predict multi-modal pose estimates, they can better model this symmetry. Interestingly, our baseline of Sample + SDF Error performed extremely poorly and only slightly better than selecting a pose estimate at random. It appears that because shape error is computed using predicted object shape, and not ground truth, it is too noisy of a signal to be directly useful ---it can disambiguate between highly probable pose-candidates but is not suitable for disambiguating between arbitrary poses. Furthermore, while pose estimation with HBEO-MDNs is slower than with HBEOs, the most expensive HBEO-MDN Posterior method can still produce multiple predictions a second; if more time is available, a higher quality estimate can be obtained by increasing the number of evaluated samples. For a fixed time budget, the Posterior version of HBEO-MDNs outperformed the Likelihood variety, even though approximating the MAP will necessitate using fewer samples than the MLE.

We also compared against two RGB-based category-level pose estimation methods, Pix3D \cite{sun2018pix3d} and Render For CNN \cite{Su_2015_ICCV} on the Pix3D dataset. All methods were trained on the ShapeNet chair category and evaluated on the $2894$ non-occluded chairs in Pix3D; Pix3D and Render for CNN used RGB images as their input while HBEOs and HBEO-MDNs were provided rendered depth-images from the same viewpoint. Because Pix3D and Render for CNN produce discretized poses, the output of HBEO and HBEO-MDN was discretized for comparison. Table \ref{table:results_table_pix3d} contains these results for multiple levels of discretization (a larger number of possible views equates to requiring more pose-estimation accuracy and the entries in the table indicate the proportion of objects with correctly estimated pose-bin). Our full method, HBEO-MDN-Posterior\footnote{HBEO-MDN variants utilized 100 samples.} achieved the best performance, along both azimuth and elevation, of all models when the number of bins was high, with HBEOs performing competitively as the bin size became larger. Interestingly, HBEOs slightly outperformed HBEO-MDN with very coarse bins (90$^\circ$) which we hypothesize is due to chair symmetry. While chairs are highly asymmetrical vertically, some variants of chairs lack arms and are thus fairly symmetrical rotationally. Because HBEOs learn to pick the average of good solutions, their predictions may be more likely to fall within 90 degrees of the true solution than HBEO-MDNs---which will tend to predict a mode of the distribution instead of the mean. This is primarily an artifact of using a sampling strategy to select a single pose instead of evaluating the entire MDN-predicted pose distribution.

\begin{table}
  \caption{Pix3D pose estimation performance --- discretized predictions (higher is better)}
  \label{table:results_table_pix3d}
  \centering
  \begin{tabular}{lccccccc}
    \toprule
     \multicolumn{8}{r}{\hphantom{1234567898765432}Azimuth \hphantom{123456789876} Elevation\hphantom{1234}} \\
    \cmidrule(r){2-5} \cmidrule(r){6-8}
    Number of Bins & 4 & 8 & 12 & 24 & 4 & 6 & 12 \\
    \midrule
Render For CNN \cite{Su_2015_ICCV} & $0.71$ & $0.63$ & $0.56$ & $0.40$ & $0.57$ & $0.56$ & $0.37$ \\
Pix3D \cite{sun2018pix3d} & $0.76$ & $0.73$ & $0.61$ & $0.49$ & $0.87$ & $0.70$ & $0.61$ \\
HBEO \cite{hbeo2018} & $\mathbf{0.87}$ & $0.76$ & $0.69$ & $0.53$ & $0.96$ & $0.91$ & $0.71$ \\
HBEO-MDN-Likelihood (Ours) & $0.78$ & $0.73$ & $0.70$ & $0.59$ & $\mathbf{0.97}$ & $\mathbf{0.93}$ & $\mathbf{0.75}$ \\
HBEO-MDN-Posterior (Ours) & $0.80$ & $\mathbf{0.76}$ & $\mathbf{0.73}$ & $\mathbf{0.62}$ & $\mathbf{0.97}$ & $\mathbf{0.93}$ & $\mathbf{0.75}$ \\
    \bottomrule
  \end{tabular}
\end{table}

\section{Conclusion}
We introduced a depth-based 3DOF category-level pose estimation algorithm, applicable to novel objects, that leverages probabilistic output and a 3D shape prediction to verify estimated pose against the observed depth-image. Our method is agnostic to the underlying generative model, provided it produces explicit 3DOF-pose and 3D-shape estimates, and we use it to extend the HBEO framework into HBEO-MDNs. Experimentally, HBEO-MDNs significantly improved on the existing state-of-the-art, providing a significant reduction in average-case pose error and incidence of catastrophic pose-estimation failure; an ablation analysis demonstrated that both the probabilistic output and estimated pose-prior contribute to this result. Our approach is fast enough for use in robotics applications, performing inference at multiple hertz, and is variable-time, allowing increased time budgets or computational power to result in improved pose-estimation.

\bibliographystyle{plainnat}
\bibliography{main}
\end{document}